\documentclass{article}

\usepackage[final]{neurips_2025}

\usepackage[utf8]{inputenc} 
\usepackage[T1]{fontenc}    
\usepackage{hyperref}       
\usepackage{url}            
\usepackage{booktabs}       
\usepackage{amsfonts}       
\usepackage{nicefrac}       
\usepackage{microtype}      
\usepackage{xcolor}         
\usepackage{amsmath}
\usepackage{cleveref}
\usepackage{graphicx}
\usepackage{float}
\usepackage{lipsum}
\usepackage{xcolor}

\definecolor{myred}{HTML}{d73027}
\definecolor{myblue}{HTML}{4575b4}

\title{Disaggregation Reveals Hidden Training Dynamics: \\ The Case of Agreement Attraction}

\author{
  James A. Michaelov \\
  MIT \\
  \texttt{jamic@mit.edu} \\
  \And
  Catherine Arnett \\
  EleutherAI \\
  \texttt{catherine@eleuther.ai} \\
}

\begin{document}

\maketitle

\begin{abstract}
Language models generally produce grammatical text, but they are more likely to make errors in certain contexts. Drawing on paradigms from psycholinguistics, we carry out a fine-grained analysis of those errors in different syntactic contexts. We demonstrate that by disaggregating over the conditions of carefully constructed datasets and comparing model performance on each over the course of training, it is possible to better understand the intermediate stages of grammatical learning in language models. Specifically, we identify distinct phases of training where language model behavior aligns with specific heuristics such as word frequency and local context rather than generalized grammatical rules. We argue that taking this approach to analyzing language model behavior more generally can serve as a powerful tool for understanding the intermediate learning phases, overall training dynamics, and the specific generalizations learned by language models.
\end{abstract}

\section{Introduction}

Though only a recent development (see, e.g., \citealp{linzen_assessing_2016,marvin_targeted_2018,wilcox_what_2018,warstadt_neural_2019,warstadt_blimp_2020,gauthier_syntaxgym_2020,hu_systematic_2020}), it is almost taken for granted today that language models tend to generate grammatical strings of text. Indeed, contemporary large language models have been argued to show linguistic competence \citep{mahowald_dissociating_2024}. But even large models such as Chinchilla have been shown to often fail at more difficult grammatical tasks \citep{lampinen_can_2024}, suggesting that rather than learning fully general rules, models may be learning more specific rules or increasingly complex heuristics. In this study, we investigate what generalizations language models \textit{do} learn by turning to two highly influential approaches from the study of human language---analyzing errors \citep[e.g.,][]{fromkin_non-anomalous_1971,garrett_analysis_1975,dell_spreading-activation_1986,bock_broken_1991} and studying changes over the course of acquisition \citep[e.g.,][]{kenney_acquisition_1972,rumelhart_parallel_1986,marcus_overregularization_1992}.

In this paper, we focus on subject-verb agreement, which is the fact that in a sentence such as \textit{the cat leaps}, the word \textit{leaps} correctly agrees with the subject \textit{cat}; while \textit{the cat leap} is grammatically incorrect. In the simplest cases, subject-verb agreement appears relatively easy to learn for language models. It is learned early in training \citep{evanson_language_2023} and at human-scale levels of training data \citep{warstadt_findings_2023,hu_findings_2024,wilcox_bigger_2025}, and can be learned even for low-resource languages \citep{jumelet_multiblimp_2025} and by traditional LSTM-RNNs \citep{linzen_assessing_2016}. However, language models appear to struggle at subject-verb agreement in more complex linguistic structures that are also difficult for humans. For example, while language models are good at predicting the correct form of the verb (V) in sentences like (\ref{eq:bb_simple}) based on the subject noun (S), their overall performance drops in cases where there is an intervening attractor noun (A) as in (\ref{eq:bb_nounpp}) \citep{marvin_targeted_2018,gulordava_colorless_2018,warstadt_blimp_2020,arehalli_neural_2020,ryu_accounting_2021,lakretz_mechanisms_2021,lampinen_can_2024}, and they are also slower to reach their peak performance on such items \citep{evanson_language_2023}. The crucial point to note, however, is that language model performance is not uniformly worse on such stimuli---like humans \citep[see, e.g.][]{bock_broken_1991,bock_regulating_1992,franck_subject-verb_2002}, they have been observed to show a higher error rate in cases where there is a mismatching attractor, i.e., in cases such as \textit{the athletes near the bike \textbf{know}/\textbf{knows}} (\citealp{arehalli_neural_2020,ryu_accounting_2021}; though see \citealp{lampinen_can_2024}). This may suggest that the models are not making their predictions on the basis of a general subject-verb agreement rule, but rather based on more specific patterns or or surface-level heuristics (for a more general discussion on language model behavior in this vein, see, e.g., \citealp{mccoy_embers_2024}). 

\begin{equation}\label{eq:bb_simple}
\begin{array}{ll}
\text{The }\left\{
\begin{array}{l}
\text{\textcolor{myred}{athlete}} \\
\text{\textcolor{myblue}{athletes}}
\end{array}
\right\}_{\text{S}}~
\text{ }\left\{
\begin{array}{l}
\text{\textcolor{myred}{knows}} \\
\text{\textcolor{myblue}{know}}
\end{array}
\right\}_{\text{V}}...
\end{array}
\end{equation}

\begin{equation}\label{eq:bb_nounpp}
\begin{array}{ll}
\text{The }\left\{
\begin{array}{l}
\text{\textcolor{myred}{athlete}} \\
\text{\textcolor{myblue}{athletes}}
\end{array}
\right\}_{\text{S}}
\text{ near the }\left\{
\begin{array}{l}
\text{\textcolor{myred}{bike}} \\
\text{\textcolor{myblue}{bikes}}
\end{array}
\right\}_{\text{A}}~\text{ }\left\{
\begin{array}{l}
\text{\textcolor{myred}{knows}} \\
\text{\textcolor{myblue}{know}}
\end{array}
\right\}_{\text{V}}...
\end{array}
\end{equation}

We carry out an exploratory analysis looking at how the language model performance on different data subsets---corresponding to different experimental manipulations of sentences of the form shown in (\ref{eq:bb_simple}) and (\ref{eq:bb_nounpp})---vary over the course of training, and how these patterns compare to those observed in when looking at the aggregate performance across all sentence types. 

\section{Method}

\paragraph{Datasets}
We use the \texttt{simple\_english} and \texttt{nounpp\_english} (i.e., subject-verb agreement with prepositional phrase attractor) subsets of the Subject-Verb Agreement task \citep{linzen_assessing_2016,marvin_targeted_2018,gulordava_colorless_2018,goldberg_assessing_2019,wolf_additional_2019,lakretz_emergence_2019,lakretz_mechanisms_2021} in BIG-bench \citep{srivastava_beyond_2023}. We also add to this the corresponding (i.e., subject-verb agreement with prepositional phrase attractor) subset of the stimuli from \citet{bock_regulating_1992}, as preprocessed by \citet{arehalli_neural_2020}; which differ from the BIG-bench stimuli in that they all only include the verb \textit{to be} (i.e., \textit{is}/\textit{are}), and that in some cases, the subject is more than one word long (e.g., \textit{the teaching assistant}). We additionally create simple agreement sentences by removing the intervening prepositional phrases from these stimuli. 

\paragraph{Models}
We use the PolyPythia suite of language models \citep{van_der_wal_polypythias_2024}. These are a set ten random seeds of each Pythia model \citep{biderman_emergent_2023} from 14M to 410M parameters, released with multiple checkpoints over the course of training. Using these models allows us to look at the training dynamics of the models while accounting for differences arising from random initialization and data shuffling. Each training step represents the same number of tokens seen (with the same tokenizer), therefore, we can compare across model sizes and random seeds.

\paragraph{Procedure}
We calculate the log-probability of the each verb following its context. We consider the model to be correct if it assigns a higher log-probability to the correct form of the verb relative to the incorrect form. For some verbs, the singular and plural forms are each represented by one token; while for others, the singular form (e.g., \textit{admires}) is represented by two tokens and the plural form (e.g., \textit{admire}) by one. We refer to these as single-token and multi-token verbs, respectively, and analyze them separately. We define the log-probability of multi-token words as the sum of their token log-probabilities (as a further analysis, we also consider normalizing by number of tokens; see \Cref{app:normalization}). We release all code in the following repository: \url{https://github.com/jmichaelov/sv-disaggregation-cognitive-interpretability}.

\section{Results}

\begin{figure}[h]
    \centering
    \includegraphics[width=\linewidth]{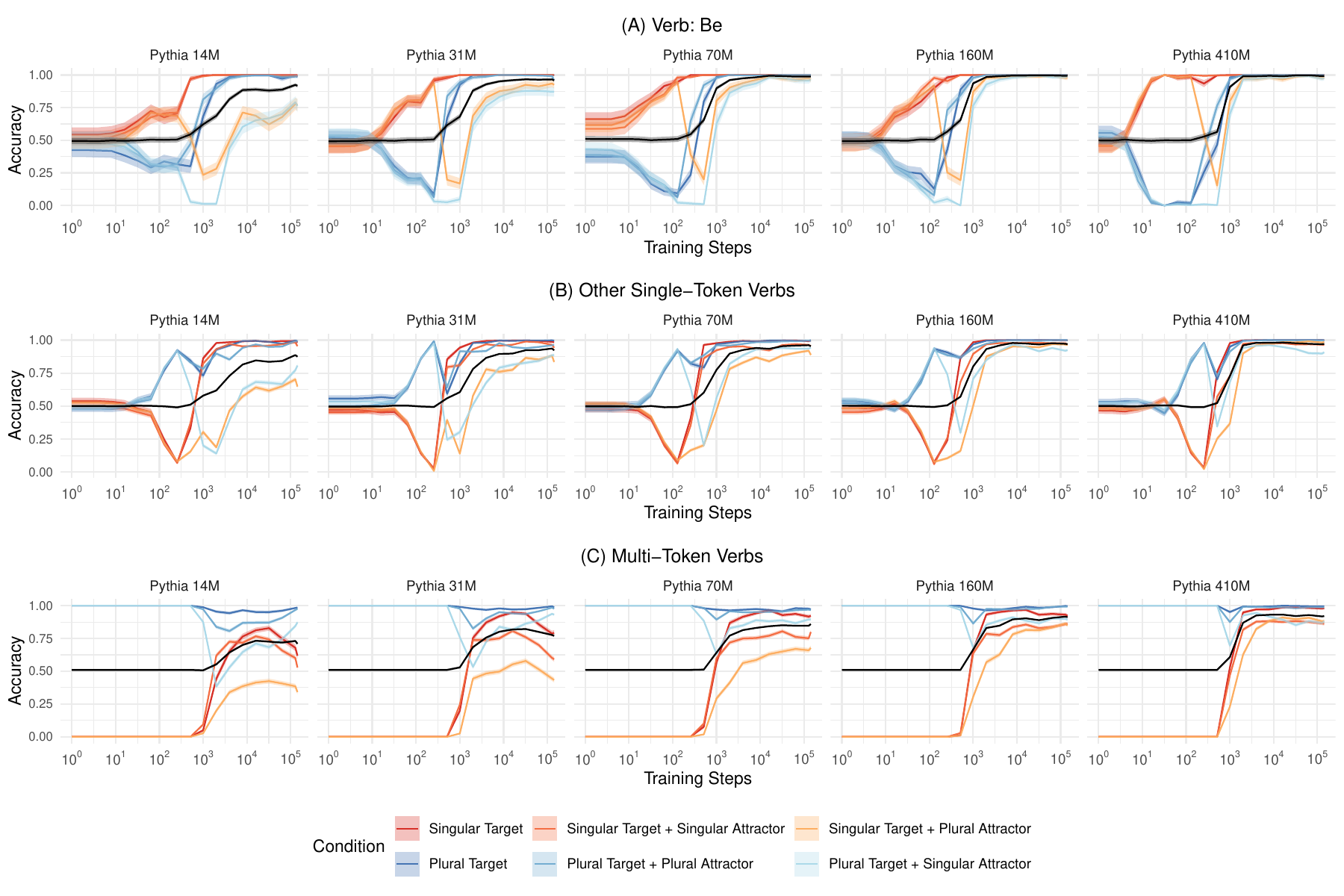}
    \caption{PolyPythia model accuracy on subject-verb agreement stimuli with (A) the verb \textit{be}, (B) all other single-token and (C) multi-token verbs. The black line represents the mean across all conditions (i.e., the aggregate score). Shading reflects 95\% confidence intervals.}
    \label{fig:unnormed}
\end{figure}

We present the results of our analysis in \Cref{fig:unnormed}. First, we consider performance as measured by the average accuracy across all conditions. In general, we see little improvement (and in some cases, a drop in accuracy) until steps 256--512, at which point we see an increase in performance, which is smaller and more gradual for the smaller models (and begins later for the Pythia 14M) and faster and more complete for the larger models.

Condition-level accuracy reveals a rather different set of patterns. For the \textit{be} verb (i.e., \textit{is} vs. \textit{are}), we initially see a high accuracy for the singular conditions and a low accuracy for the plural conditions; i.e., the model is virtually always predicting the word \textit{is} to be more likely than \textit{are}. Then, around steps 128-512, we see a sharp increase in accuracy for the plural and plural-with-plural-attractor conditions, but not the plural-with-singular-attractor condition; and we see a corresponding sharp decrease in singular-with-plural-attractor condition---that is, we see the agreement attractor effect. After this, we see an overall increase in performance for all conditions. For the other verbs, we see the reverse pattern---a preference for the plural (e.g., \textit{admire}) over the singular (e.g., \textit{admires}), followed by a drop in performance of the plural-with-singular-attractor condition and an increase in the singular and singular-with-singular-attractor conditions, and finally, an overall increase. Disaggregating the results for each verb (see \Cref{app:verb_level}), we see that while most of these patterns are present for each verb and each model, there is some variation---for example, the preference for \textit{stimulate} over \textit{stimulates} is much smaller than for the other verbs. Additionally, in some cases, there appears to be a brief reversal of the singular vs. plural preference at step 512 (for example, with \textit{observe}), though this quickly reverses. There is also some variation by random seed (\Cref{app:seed_level}). In general, the patterns are less stable overall in the smaller models.

For multi-token verbs, we see overall the same pattern: an initial (and more immediate) preference for (one-token) plural over (two-token) singular verbs, followed by an increase in accuracy for verbs in the singular and singular-with-singular-attractor conditions and a corresponding decrease in accuracy in the plural-with-singular-attractor condition, followed by overall improvement on all conditions. However, these patterns generally occur later in training, and the performance decrease is smaller.

\section{Discussion}
In this study, we show that aggregated metrics of performance may hide interpretable patterns in the trajectory of language models' grammatical knowledge over the course of training. We see evidence of clear systematic patterns at both the condition level and based on verb tokenization. In contrast to the overall slow and gradual increase in performance, disaggregation reveals rapid (and often non-monotonic) changes in the behavior that underlie this and that begin far earlier. Initially, models learn to assign a higher probability to the more frequent form of the verb. In the case of \textit{be}, \textit{is} is more frequent than \textit{are}; and in all other cases, the plural form of the verb (as the bare form of the verb) is more frequent in the training corpus (see \Cref{app:v_freqs}). Next, the models appear to become sensitive to the preceding word: we see a sharp improvement in performance on the simple condition of the less frequent verb form, as well as the condition with the matching attractor. There is also a decrease in performance on the mismatched attractor condition for the more frequent verb form; i.e., we see a strong effect of the attractors on performance. Finally, performance continues to improve until the end of training. A possible explanation for the first two phases is the finding that over the course of training, transformers overfit their predictions to token unigram probability (i.e., frequency), then bigram probability, then trigram probability, and so on \citep[as described in, e.g.,][]{chang_characterizing_2024}. This may also explain the later-occurring preference for singular verbs following singular nouns (targets or attractors) in multi-token verbs relative to single-token verbs. For single-token verbs, a sensitivity to the number of the preceding noun only requires taking into account the previous token (i.e., bigram-like behavior), but for the second token of a multi-token verb (which only occurs for singular forms of the verb), it requires taking into account the previous two (trigram-like behavior). Whether the models are displaying strictly $n$-gram-like behavior or a more general ability to make predictions based on an increasingly long context is a question for future work.

In either case, these findings highlight the importance of considering simple heuristics in the analysis of language model behavior. On the one hand, if a task is solvable based on bigram statistics, it may indicate that the task may not have sufficient construct validity. Indeed, the fact that a 5-gram can score well above chance on a number of BLiMP subtasks (see \citealp{warstadt_blimp_2020}) could mean models do not have to learn generalized grammatical rules to solve them. On the other hand, if it can in fact support the generation of grammatical text in the majority of cases (given the pressure for shorter dependencies in natural language; see, e.g., \citealp{futrell_dependency_2020}), it may instead be useful to think of $n$-gram-like behavior as an explanation for models' observed grammatical performance rather than as a confound.

Our results provide another piece of evidence in the debate about whether learning in language models is sudden \citep[e.g.,][]{wei_emergent_2022,olsson_-context_2022,power_grokking_2022,chen_sudden_2024,aoyama_language_2025} or gradual \citep{schaeffer_are_2023}. In line with \citet{kangaslahti_hidden_2025}, our results provide evidence that at least in some cases learning involves multiple `hidden breakthroughs' that can underlie more apparently gradual learning trajectories seen on aggregate benchmarks. We also see that while such `hidden breakthroughs' can lead to substantially better performance on some data subsets, they can also lead to substantially poorer performance on others, and that both of these an be relatively invisible when looking only at the aggregate score. 

Taken together, our results highlight how a targeted analysis of data subsets over the course of training can provide interpretable explanations for language model behavior even without additional mechanistic analyses. With grammatical benchmarks, especially those based on comparing performance on minimal pairs \citep[e.g.,][]{linzen_assessing_2016,marvin_targeted_2018,gulordava_colorless_2018,warstadt_blimp_2020,gauthier_syntaxgym_2020} the different versions of each sentence are interpretable by their nature. In many cases, they are drawn from---or based on---previous psycholinguistic research where these experimental manipulations are explicitly designed to highlight specific interpretable differences in human behavior. Thus, using the disaggregated versions of such datasets (as in, e.g., \citealp{arehalli_neural_2020,ryu_accounting_2021}) allows comparisons across experimental conditions, and is thus inherently interpretable, as it is in the original human studies \citep[see, e.g.,][]{bock_regulating_1992}. We move beyond such analyses at a single snapshot of model behavior by adding the dimension of time. As with humans, looking at patterns of behavior at various stages over the course of training can be informative. Specifically, we can see how performance at different conditions relative to each other varies over time, allowing us to characterize behavior as falling into different interpretable phases.

Beyond grammatical and other constructed benchmarks, such disaggregation based on \textit{a priori} theoretical constructs is likely to be more difficult; however, such theory-driven work is likely to be key in furthering our understanding of how language models come to have the capabilities they display. Additionally, bottom-up approaches for identifying meaningfully-different subsets exist (see, e.g., \citealp{kangaslahti_hidden_2025}), and subsets identified this would could then be further characterized to gain a better understanding of model behavior and how it changes during training.

\section*{Limitations}

This work has several limitations. First, we only investigate language model performance on English subject-verb agreement, and only consider attractors occurring within prepositional phrases. Investigating whether the trends we observe generalize beyond this limited scope would be a valuable direction for future work. Additionally, our study only uses the PolyPythia suite because we are not aware of any set of pretrained language models of multiple sizes with multiple random seeds that have a similar number of checkpoints for the early stages of training that are relevant to the present study. Finally, our work is explanatory. While this study has demonstrated the utility of disaggregation and provided a new analysis of grammatical learning in language models, it may be premature to draw any strong conclusions about the latter without further confirmatory analyses based on more specific predictions.

\section{Conclusions}
Our results suggest that the learning of grammatical rules such as subject-verb agreement by language models is neither sudden nor gradual---instead, it proceeds in a sequence of `hidden breakthroughs' \citep{kangaslahti_hidden_2025} that roughly correspond to the language model learning agreement patterns with increasingly longer dependencies. Thus, substantial insights into model behavior can be gained by analyzing performance at the level of specific subsets over the course of training and by comparing across these subsets. Indeed, our results demonstrate one can observe complex training dynamics even when analyzing behavior using a simplistic evaluation metric such as accuracy.

\section*{Acknowledgments}
We would like to thank the members of the Computational Psycholinguistics Laboratory at MIT and the Language and Cognition Laboratory at UCSD for their valuable advice and discussion. James Michaelov was supported by a grant from the Andrew W. Mellon foundation (\#2210-13947) during the writing of this paper.

\bibliography{references}
\bibliographystyle{apalike}

\appendix

\section{Verb Frequencies}
\label{app:v_freqs}

We provide the frequency of each verb form in The Pile \citep{gao_pile_2020}, the corpus on which all language models were trained. Frequencies were calculated using the \textit{infini-gram} web interface \citep{liu_infini-gram_2024}.

\begin{table}[H]
\centering
\caption{Frequency of each form of each verb analyzed in The Pile.}
\begin{tabular}{@{}ccccccc@{}}
\toprule
\textbf{}     & \textbf{} & \multicolumn{2}{c}{\textbf{Singular}} & \textbf{} & \multicolumn{2}{c}{\textbf{Plural}} \\ \cmidrule(lr){3-4} \cmidrule(l){6-7} 
\textbf{Verb} & \textbf{} & \textbf{Word}   & \textbf{Frequency}  & \textbf{} & \textbf{Word}  & \textbf{Frequency} \\ \midrule
be            &           & is              & 2,055,643,528       &           & are            & 816,249,141        \\
admire        &           & admires         & 97,112              &           & admire         & 868,285            \\
approve       &           & approves        & 233,065             &           & approve        & 1,522,021          \\
avoid         &           & avoids          & 878,590             &           & avoid          & 19,190,343         \\
confuse       &           & confuses        & 159,992             &           & confuse        & 637,652            \\
criticize     &           & criticizes      & 138,875             &           & criticize      & 521,911            \\
discourage    &           & discourages     & 102,410             &           & discourage     & 556,694            \\
encourage     &           & encourages      & 1,059,654           &           & encourage      & 4,190,741          \\
engage        &           & engages         & 636,706             &           & engage         & 4,491,021          \\
greet         &           & greets          & 141,839             &           & greet          & 1,886,071          \\
inspire       &           & inspires        & 358,885             &           & inspire        & 1,206,643          \\
know          &           & knows           & 11,077,961          &           & know           & 130,967,397        \\
observe       &           & observes        & 629,563             &           & observe        & 5,397,406          \\
remember      &           & remembers       & 937,143             &           & remember       & 16,943,328         \\
stimulate     &           & stimulates      & 705,576             &           & stimulate      & 1,522,151          \\
understand    &           & understands     & 1,385,876           &           & understand     & 30,822,538         \\ \bottomrule
\end{tabular}

\end{table}

\clearpage

\section{Verb-Level Plots}
\label{app:verb_level}
\begin{figure}[h]
    \centering
    \includegraphics[width=\linewidth]{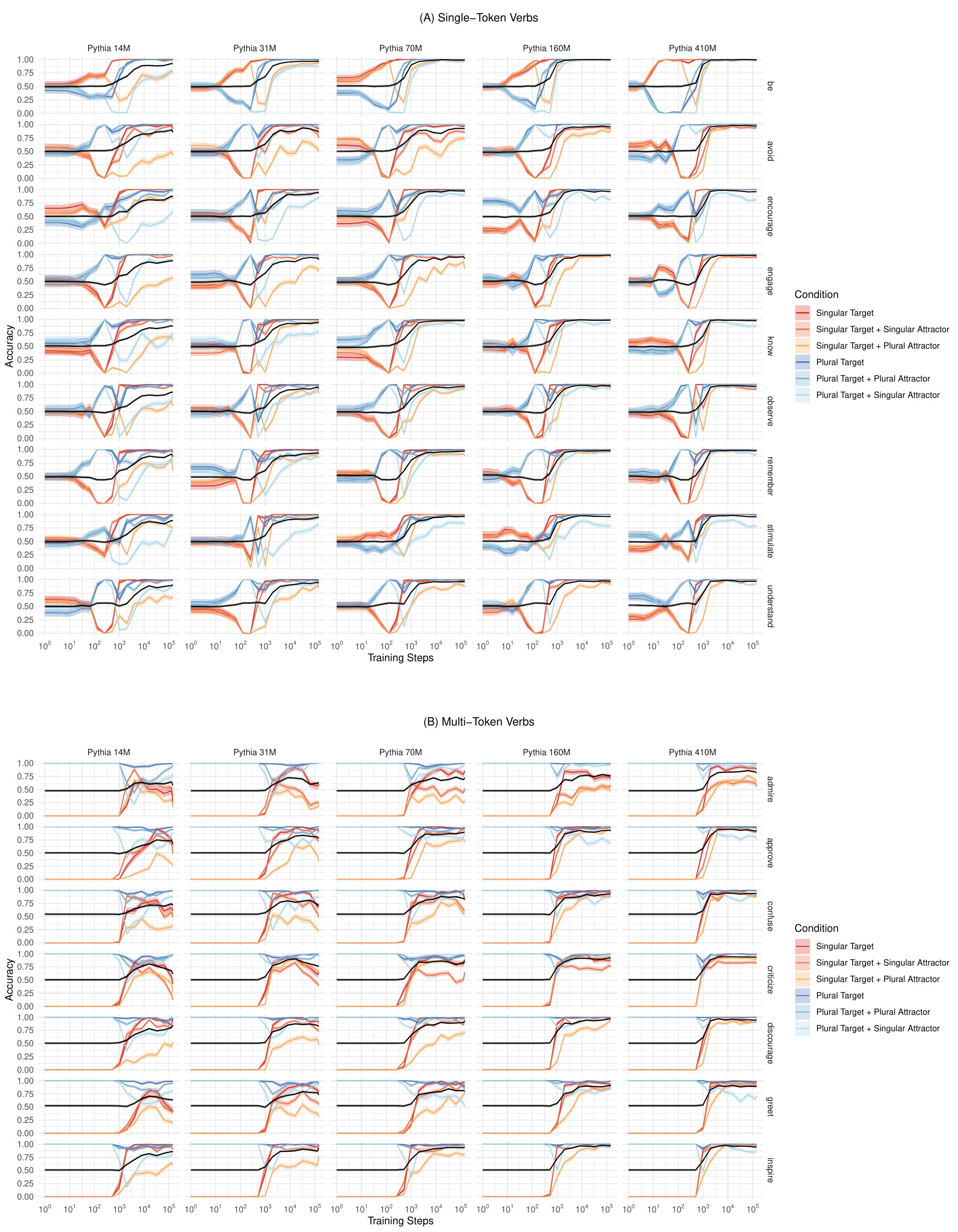}
    \caption{PolyPythia model accuracy on subject-verb agreement stimuli with (A) single-token and (B) multi-token verbs. The black line represents the mean across all conditions (i.e., the aggregate score). Shading reflects 95\% confidence intervals.}
    \label{fig:unnormed_verb}
\end{figure}

We provide verb-level plots for each single-token and multi-token verb in \Cref{fig:unnormed_verb}.

\section{Seed-Level Plots}
\label{app:seed_level}
\begin{figure}[h]
    \centering
    \includegraphics[width=\linewidth]{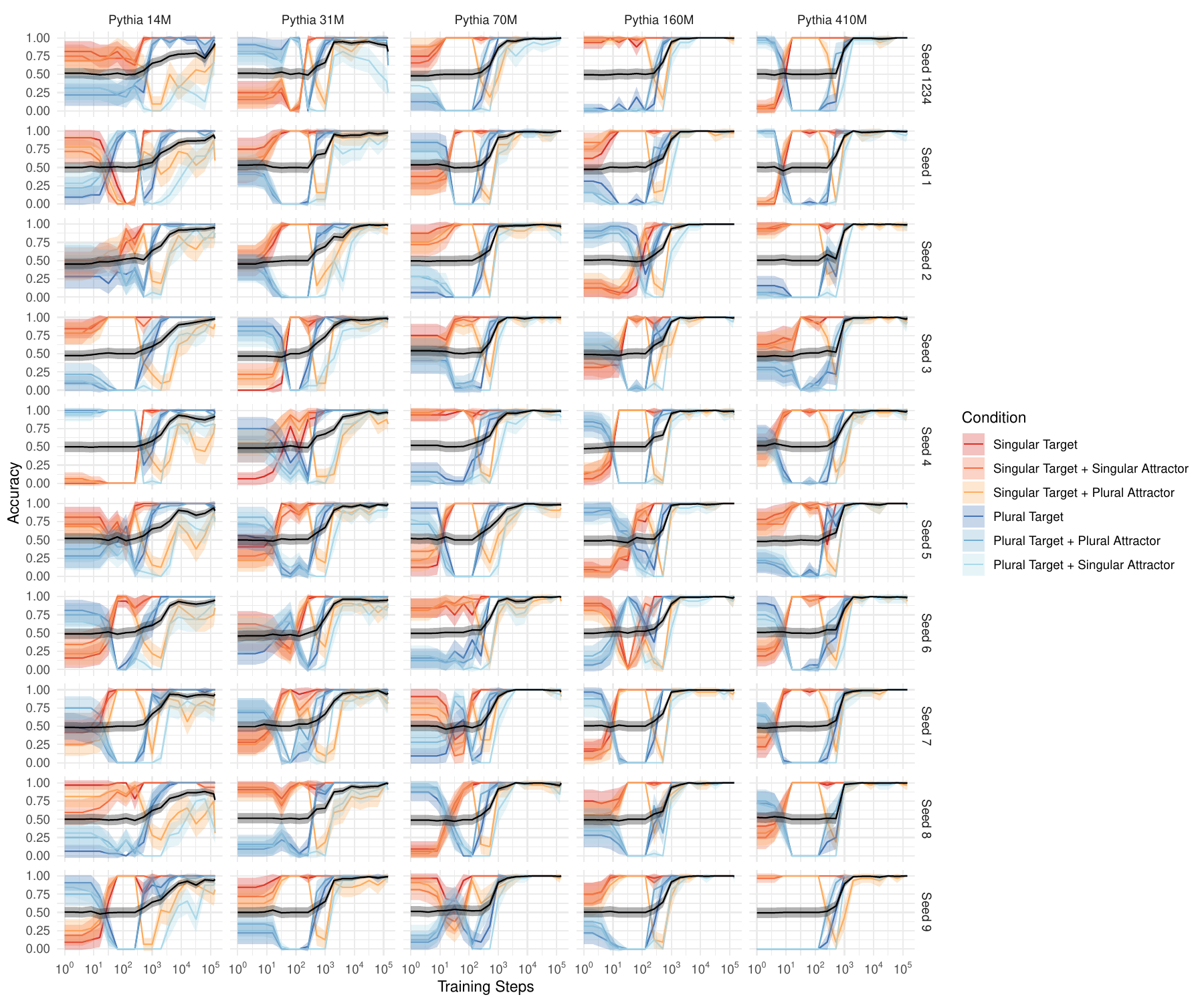}
    \caption{PolyPythia model accuracy on subject-verb agreement stimuli for the verb \textit{be}. The black line represents the mean across all conditions (i.e., the aggregate score). Shading reflects 95\% confidence intervals.}
    \label{fig:seed_be}
\end{figure}

\begin{figure}[h]
    \centering
    \includegraphics[width=\linewidth]{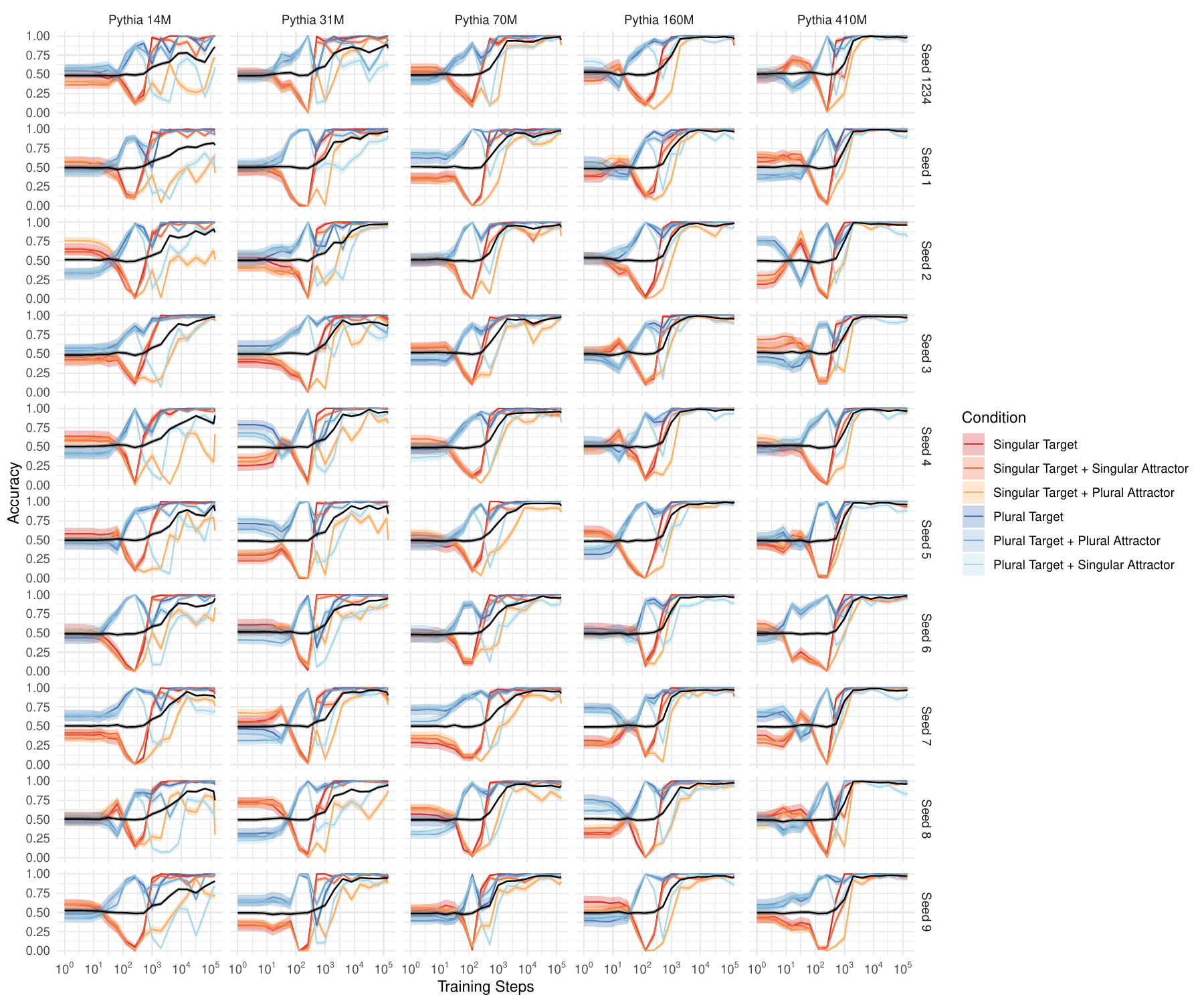}
    \caption{PolyPythia model accuracy on subject-verb agreement stimuli for single-token verbs. The black line represents the mean across all conditions (i.e., the aggregate score). Shading reflects 95\% confidence intervals.}
    \label{fig:seed_st}
\end{figure}

\begin{figure}[h]
    \centering
    \includegraphics[width=\linewidth]{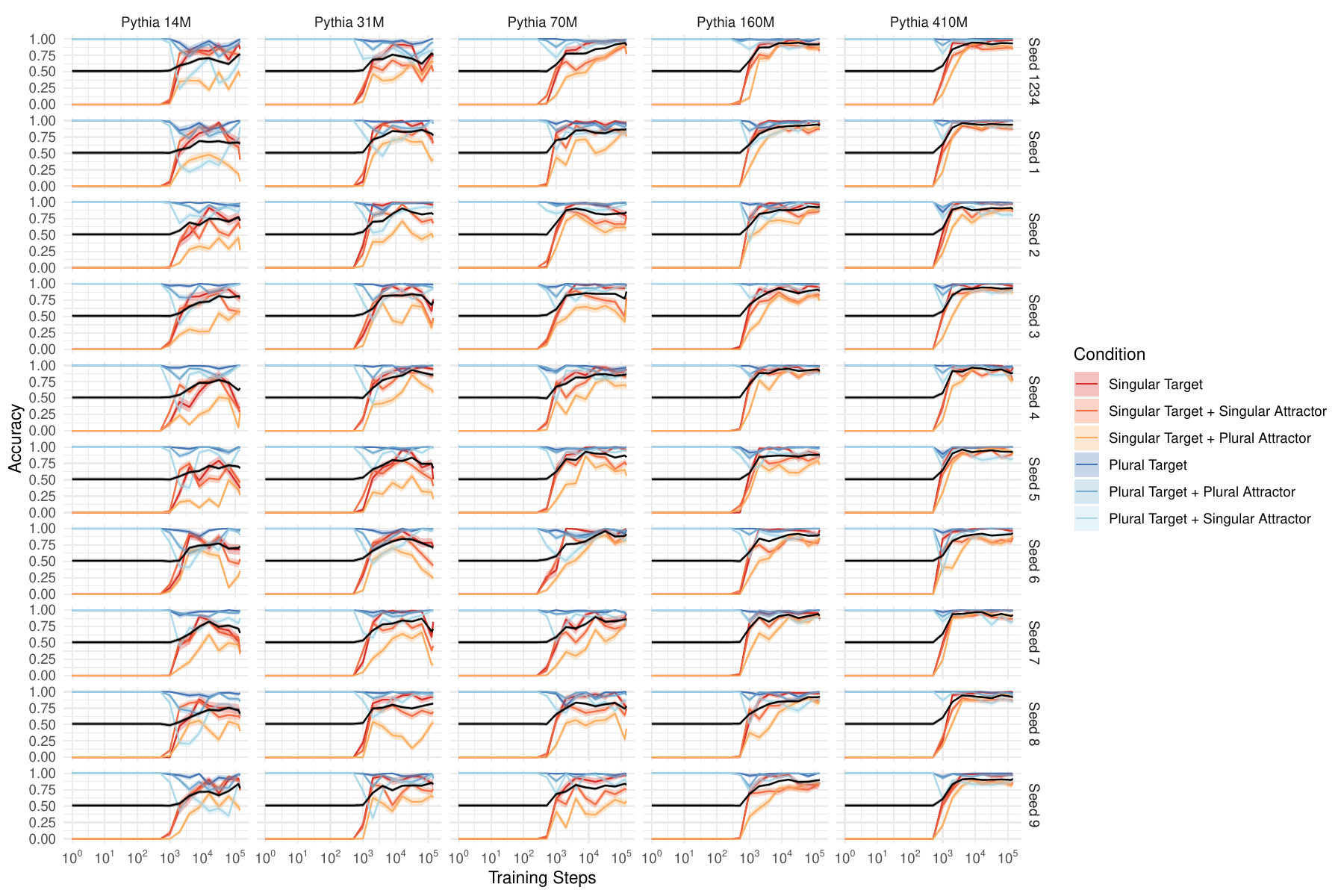}
    \caption{PolyPythia model accuracy on subject-verb agreement stimuli for multi-token verbs. The black line represents the mean across all conditions (i.e., the aggregate score). Shading reflects 95\% confidence intervals.}
    \label{fig:seed_mt}
\end{figure}

We provide seed-level plots for each \textit{be} (\Cref{fig:seed_be}), other single-token verbs (\Cref{fig:seed_st}), and multi-token verbs (\Cref{fig:seed_mt}).

\clearpage

\section{Token Normalization}
\label{app:normalization}
\begin{figure}[h]
    \centering
    \includegraphics[width=\linewidth]{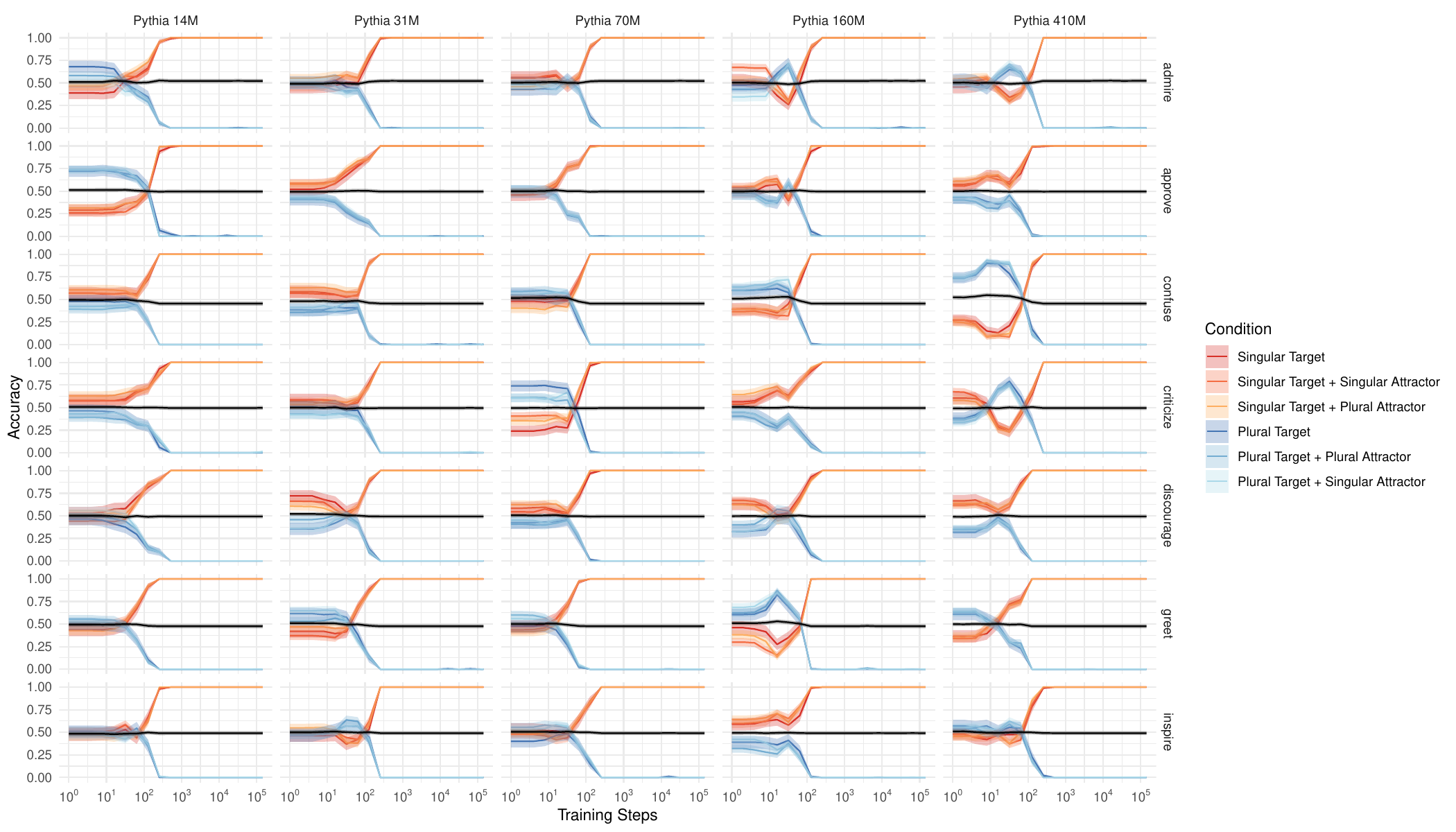}
    \caption{PolyPythia model accuracy (based on normalized log-probability) on subject-verb agreement stimuli with multi-token verbs. The black line represents the mean across all conditions (i.e., the aggregate score). The shaded areas reflect 95\% confidence intervals. }
    \label{fig:normed_mt}
\end{figure}

When using language model log-probability as a way to assess their grammatical capabilities, an open question is how to score each sentence \citep{lau_grammaticality_2017}---most commonly, whether to take the sum \citep{hu_prompting_2023} or the mean \citep{jumelet_multiblimp_2025} log-probability of each token in the relevant region (i.e., the word or sentence for which log-probability is being calculated). The former gives the `true' log-probability assigned to the sequence by the model, while the latter normalizes to account for the fact that all else being equal, a longer token sequence will be a sum over a larger number of log-probabilities, and thus likely to have a lower log-probability in total. We carry out both analyses for the multi-token verbs (there is by definition no difference for single-token verbs). As can be seen in \Cref{fig:normed_mt}, normalizing by the number of tokens (i.e., taking the mean log-probability) appears to be too strong---after the early stages of training, the mean log-probability of two-token singular verb is always higher than the log-probability of the one-token plural form.

\end{document}